\documentclass{bmvc2k}

\usepackage{colortbl}
\usepackage{soul}
\usepackage{url}
\usepackage[utf8]{inputenc}
\usepackage[small]{caption}
\usepackage{graphicx}
\usepackage{amsmath}
\usepackage{amsthm}
\usepackage{amssymb}
\usepackage{booktabs}
\usepackage{arydshln}
 \usepackage{mathtools}
 \usepackage{comment}
\usepackage{svg}
\usepackage{multirow}
\usepackage{subcaption}
\usepackage{todonotes}
\usepackage{float}
\usepackage{dblfloatfix}
\usepackage{algorithm}
\usepackage{algorithmic}
\usepackage{newfloat}
\usepackage{listings}
\usepackage{diagbox} 
\usepackage{array}   
\usepackage{minibox}
\usepackage{tikz}

\title{Generalizing Teacher Networks for Effective Knowledge Distillation Across Student Architectures}

\addauthor{Kuluhan Binici}{kuluhan@comp.nus.edu.sg}{1}
\addauthor{Weiming Wu}{weiming.wu@u.nus.edu}{1}
\addauthor{Tulika Mitra}{tulika@comp.nus.edu.sg}{1}

\addinstitution{
 School of Computing\\
 National University of Singapore\\
 Singapore
}

\runninghead{Binici, Wu, Mitra}{GTN}

\begin{document}

\maketitle

\begin{abstract}
Knowledge distillation (KD) is a model compression method that entails training a compact student model to emulate the performance of a more complex teacher model. However, the architectural capacity gap between the two models limits the effectiveness of knowledge transfer. Addressing this issue, previous works focused on customizing teacher-student pairs to improve compatibility, a computationally expensive process that needs to be repeated every time either model changes. Hence, these methods are impractical when a teacher model has to be compressed into different student models for deployment on multiple hardware devices with distinct resource constraints. In this work, we propose {\em Generic Teacher Network (GTN)}, a one-off KD-aware training to create a generic teacher capable of effectively transferring knowledge to any student model sampled from a given finite pool of architectures. To this end, we represent the student pool as a weight-sharing supernet and condition our generic teacher to align with the capacities of various student architectures sampled from this supernet. Experimental evaluation shows that our method both improves overall KD effectiveness and amortizes the minimal additional training cost of the generic teacher across students in the pool.
\end{abstract}

\section{Introduction}
\label{sec:intro}
In practical scenarios, such as deploying AI solutions in mobile devices, IoT devices, and embedded systems, there is a vast spectrum of computational capabilities and memory limitations. These differences necessitate the use of neural network models of varying sizes and complexities, tailored to the specific resources available on each device. Selecting the most appropriate model is challenging as high-performing models are often resource-demanding, while the resource-efficient ones often cannot achieve high performance. To break this tradeoff, a popular approach is reducing the resource footprint of large pre-trained models, without sacrificing significant performance through model compression. \textit{Knowledge distillation} (KD) \cite{hinton2015distilling} is a prominent model compression technique that involves transferring the predictive behavior acquired by a large pre-trained model, known as the ``teacher'', to a compact ``student'' architecture. However, not every neural network effectively benefits from KD due to the potential capacity mismatch with the teacher model. 
\par
Addressing this issue, some works explored methods for bridging such capacity gap at distillation time \cite{zhu2021sckd}. These typically involve controlling the influence of the teacher model throughout the training process of the student to prevent adverse impacts on accuracy. However, despite their user-friendly nature, as they do not require modifications to the teacher model, their effectiveness is constrained, as evidenced by our experiments. An alternative approach is specializing a given teacher architecture to be conducive for effective knowledge transfer to a particular reference student \cite{park2021sftn}. This is achieved by conditioning the training process of the given teacher, which involves optimizing a snapshot of the reference student model jointly with the teacher on the target dataset. Throughout this process, the teacher is additionally constrained to match the outputs of the student snapshot. This causes the teacher to converge to a function that the student can easily approximate within its capacity later during the KD stage. However, the specialization procedure has to be repeated between every teacher-student pair as teacher models tailored to a particular student might be ineffective or even detrimental for other students, as highlighted in our experiments. This need for iterative repetition upscales the time cost, rendering the method impractical when dealing with scenarios with multiple students requiring KD. This creates a tradeoff between performance and scalability.
\par
\begin{figure}[!t]
\centering
\includegraphics[width=0.8\textwidth]{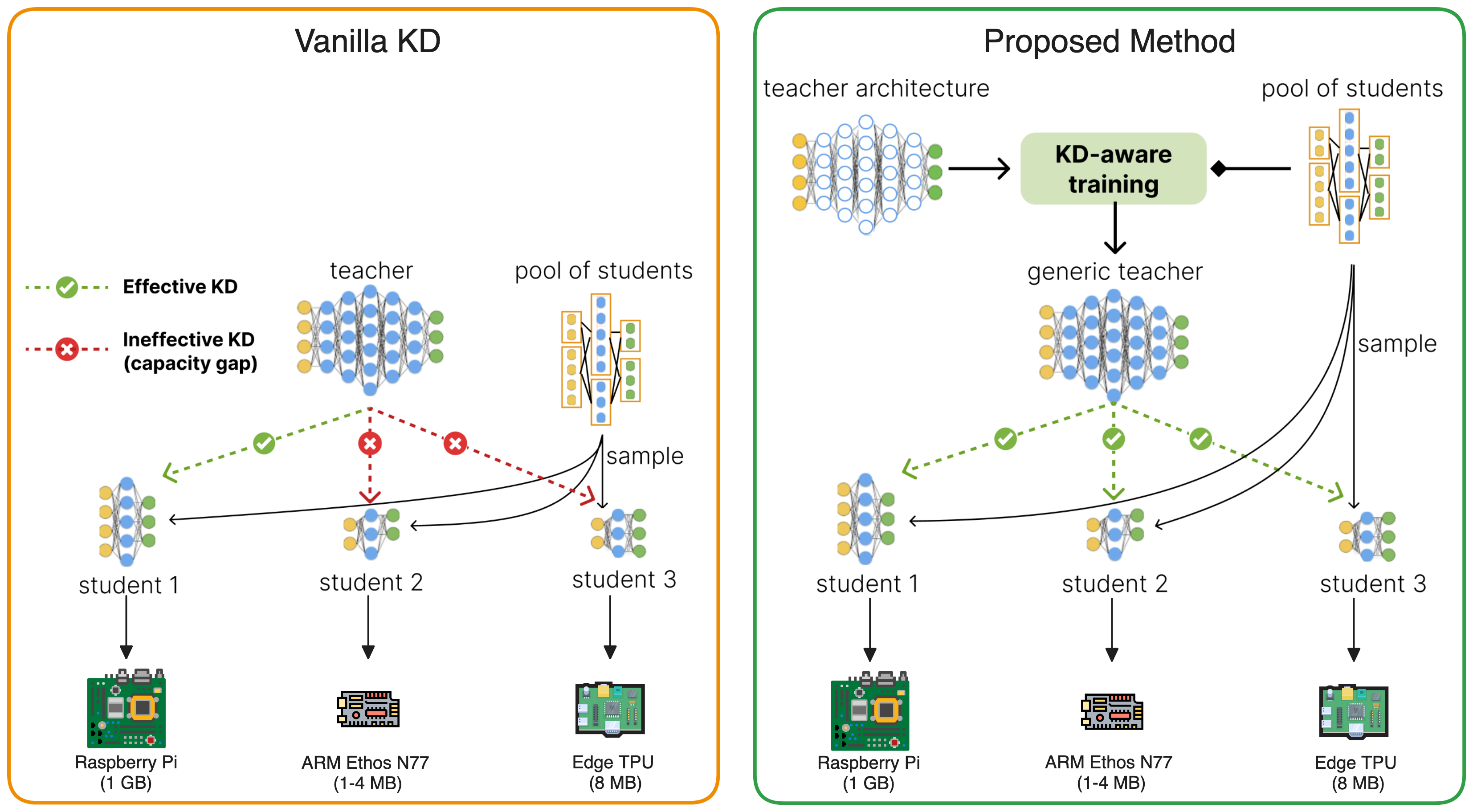}
\vspace{5pt}
\caption{Illustration of the capacity gap problem in KD and the motivation behind our proposed generic teacher approach.}
\label{fig:intro_fig}
\vspace{-15pt}
\end{figure}
In this work, we present a one-off KD-aware teacher training method that trains a generic teacher model with consideration to the capacities of all student models contained in a given finite pool of architectures, as illustrated in Figure \ref{fig:intro_fig}. This eliminates the aforementioned performance vs. scalability tradeoff by limiting the time cost to the one-off training time of the generic teacher. Our method involves partially training reference student models sampled from the pool, to regularize the output predictions of the teacher. To avoid significant time costs, we allow parameter sharing among these reference student models. For this, we configure the pool as an over-parameterized supernet architecture \cite{liu2018darts}, containing a diverse range of neural network blocks that can be connected in various ways to represent different neural network architectures. At each training iteration of the teacher model, the supernet model is reconfigured to represent a different reference student model by selecting a single NN block at each supernet layer. Therefore, parameters are shared across all students containing the same NN blocks within their architectures. By reconfiguring the supernet at each iteration, the teacher model is regularized based on the capacity of a different student model. 
\par
Experimental evaluation of our method reveals consistent improvements in the KD accuracy of students sampled from the pool of architectures, resolving the limited effectiveness of specialized teachers in generalizing to different students. While providing such flexibility, our approach maintains a constant time overhead which is equivalent to training 2-3 teacher models specialized for different students. Furthermore, as an extended evaluation, we explore the use of students selected through Neural Architecture Search (NAS) rather than manual selection from the given pool. The results of these experiments indicate the viability of our approach in NAS scenarios to boost the performance of various student models customized for different deployment platforms.
\vspace{-10pt}
\section{Related Work}
\subsection{Knowledge Distillation (KD)}
Knowledge Distillation (KD) \cite{hinton2015distilling} is a technique that involves training a compact neural network, referred to as the "student" model, to approximate the decision-making capabilities of a more complex one known as the "teacher." Typically, the student model is trained to match the logit scores or softmax probabilities of the teacher model \cite{romero2014fitnets}. However, alternative approaches, such as employing activation maps or attention scores, have also been explored \cite{zagoruyko2016paying} for this purpose. Recently Zhao et al. \cite{zhao2022decoupled} identified that the skewness of the predictive distribution of complex teachers towards target classes limits the transfer of useful knowledge from non-target classes. Introducing Decoupled Knowledge Distillation (DKD), they separate knowledge transfer from target and non-target classes. DKD dynamically adjusts the weighting of both learning objectives to ensure that the impact of non-target class-related information is not overlooked.
\vspace{-10pt}
\subsection{Teacher-Student Capacity Gap in KD}
Research by Cho et al. \cite{cho2019eff}, Mirzadeh et al. \cite{mirzadeh2020teacherassistant}, and Menon et al. \cite{menon2021statistical} challenges the notion that more accurate teachers always enhance student learning, highlighting that mismatches in model capacities can hinder KD. Liu et al. \cite{liu2020pearls} further suggest that different student models perform better with different teachers, indicating that compatibility between teacher and student models affects KD effectiveness. Zhu et al. \cite{zhu2021sckd} introduced Student-customized Knowledge Distillation (SCKD) to mitigate issues arising from the capacity gap by adapting knowledge transfer based on gradient orientations, although it does not alter the teacher's teaching abilities.
\par
To address the capacity gap more effectively, SFTN \cite{park2021sftn} customizes the teacher model for specific students, enhancing KD outcomes for those students but potentially harming others. This customization is not scalable when multiple students with different resource needs must be accommodated, leading to significant time costs.
\vspace{-10pt}
\subsection{Neural Architecture Search \& Supernet Architectures}
Neural Architecture Search (NAS) automates the design of neural architectures, often outperforming manual designs for various tasks \cite{gou2021survey, elsken2019survey}. Initial approaches utilized genetic algorithms \cite{real2019evo} or reinforcement learning \cite{zoph2016neural}, which are time-consuming due to the need to train and evaluate many models. To enhance efficiency, DARTS \cite{liu2018darts} introduced a differentiable search strategy within a weight-sharing "supernet", allowing gradient-based optimization to find optimal models. However, this method significantly increases memory consumption. ProxylessNAS \cite{cai2018proxyless} addresses this by binarizing paths in the supernet, reducing memory demands by loading only one path at a time during the search. Our method's supernet configuration is inspired by the design used in ProxylessNAS.
\vspace{-10pt}
\section{Method}
To obtain our generic teacher we condition its training process with consideration to the capacities of various reference students drawn from the given set. For such conditioning, we use the SFTN algorithm. As this would involve training snapshots of each of these reference students from scratch, we allow weight-sharing among them to avoid excessive time costs. 
This is achieved by using a supernet architecture that contains all possible candidate operations required to build any student model from the pool. 
We name our method as \textit{Generic Teacher Networks} (GTN). Technical details are discussed in the following sections.
\subsection{Conditioning the Teacher Based on the Capacity of a Reference Student Architecture}
\begin{figure*}[!t]
  \centering
\includegraphics[width=0.85\linewidth]{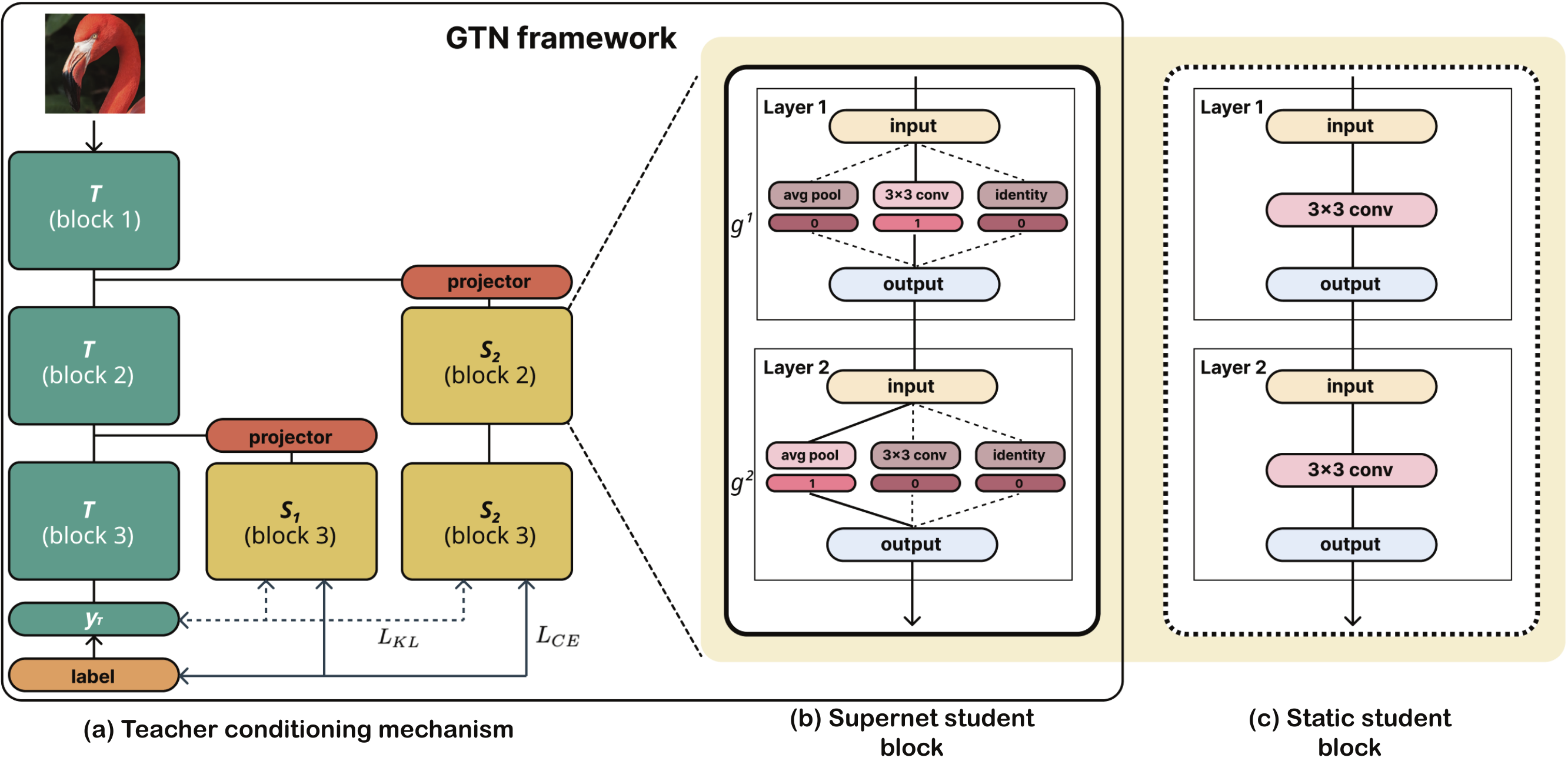} 
    \vspace{5pt}
   \caption{Overview of our GTN framework. (a) Teacher model is regularized based on the capacity of a reference student. (b) Supernet blocks allow the architecture of the student branches to be reconfigured, exposing the teacher to various reference students for regularization. (c) Static student blocks used to train specialised teachers.}
   \label{fig:framework}
   \vspace{-10pt}
\end{figure*}

This process aims to achieve two goals: (i) training a teacher model until convergence to a function that yields high accuracy, (ii) constraining the set of functions that the teacher can converge to, based on the reference students' capacity. The first target can be simply achieved by minimizing the cross-entropy between the predictions of the teacher and the ground truth labels. As for meeting the second objective, we make use of SFTN approach. First, the reference student model is partitioned into a number of blocks, and various combinations of these blocks are grafted on top of teacher blocks as in Figure \ref{fig:framework} (a). Later during training, the outputs of the teacher are forced to match those of each of these grafted student branches. Therefore the final optimization objective is minimizing the loss function in Eq. \ref{eq:l_SFTN}.
\begin{equation}
    L_{CT} = \frac{1}{n}\sum_{i=1}^{n}\left(\underbrace{\vphantom{\frac{z_t}{T}}y_{gt}\log(\hat{y}_{s_{i}})}_{L_{CE_{S}}} + \alpha \underbrace{T^{2}D_{KL}\left(\frac{z_t}{T} \middle\| \frac{z_{s_{i}}}{T}\right)}_{L_{KL}}\right) + \underbrace{\vphantom{\frac{z_t}{T}}y_{gt}\log(\hat{y}_{t})}_{L_{CE_{T}}}
\label{eq:l_SFTN}
\end{equation}
In Eq. \ref{eq:l_SFTN} $n$ denotes the number of student branches, while $\hat{y_{t}}$ and $\hat{y_{s_{i}}}$ represent the predictive outputs of the teacher and student branches indexed with $i$ respectively. $\alpha$ is a coefficient that adjusts the weighting of the $L_{KL}$ with respect to other terms. The loss term is composed of three sub-terms, namely $L_{CE_{S}}$, $L_{KL}$ and $L_{CE_{T}}$. The first and the last terms are the cross-entropy losses from the student and teacher branches calculated using the ground truth labels $y_{gt}$. Moreover, the second term, $L_{KL}$, is the KL-divergence between the re-scaled logits of the teacher and student ($z_t$ and $z_{s_{i}}$) by temperature $T$. Minimizing this term constrains the outputs of the teacher to match those of the student branches.
\subsection{GTN Training Using a Supernet as Reference Architecture}
The most straightforward way for conditioning the teacher using various reference students, would be extending the amount of grafted student branches with blocks from different architectures. However, this would increase the memory footprint significantly and render the training process infeasible on many hardware systems. Moreover, optimizing such a large number of student branches from scratch along with the teacher would consume a significant amount of time.
Instead, we propose configuring a single reference architecture that embodies all student models in the given pool, and allows weight sharing among them. For this purpose we use a supernet architecture consisting of multiple alternate paths at every layer, with each path containing a different candidate operation that maps the inputs to the outputs, as shown in Figure \ref{fig:framework} (b). Operations that are commonly shared among different student models at corresponding layers, use the same trainable parameters. This prevents excessive time costs due to re-training the same set of parameters for each student. To condition the teacher with this supernet as the reference architecture, we again partition it into blocks and create the student branches. Later during subsequent training, we employ path binarization to reduce the memory requirement, as done in Cai et al. \cite{cai2018proxyless}. This technique trains a supernet stochastically, by sampling and updating a single operation path from every layer at each iteration. Therefore, at any time, the memory occupation is limited to the size of a single sub-network contained within the supernet. In the context of our GTN framework, these sampled sub-networks correspond to individual student models from the given pool. As sampling operations randomly can cause the teacher model to be regularized by aribtrarily bad student models and damage the conditioning outcome, we instead sample them from a trainable probability distribution. For this, we use a multinomial probability distribution $p_{\phi} = softmax(\phi)$ parameterized by a single trainable random variable $\phi$. To activate the sampled operation paths and deactivate the rest, we use binary gates. For $k$-many candidate operations, we use the same amount of binary gates, with each gate $g^{j}$ controlling whether the $j^{th}$ operation will be activated or not. The index of the active operation is denoted as $j^{act}$.
\begin{equation}
    g^j = 
    \begin{cases}
        1, & \text{if } j = j^{act} \sim p_{\phi}\\
        0, & \text{otherwise }
    \end{cases}
    \label{eq:gate}
\end{equation}
The random variable $\phi$ has $k$-many possible outcomes (one for each operation) and is learned during the GTN training process by minimizing $L_{\phi}$ loss given in Eq. \ref{eq:arch_loss}. 
\begin{equation}
    L_{\phi} = L_{CE_{S}} - \alpha L_{KL}
    \label{eq:arch_loss}
\end{equation}
Different from $L_{CT}$, $L_{KL}$ appears with a negative $\alpha$ coefficient in $L_{\phi}$ This is because, ideally, the gate parameters should be updated in a way that would motivate an extensive exploration of candidate operations at each training step. While sampled student parameters $\theta$ are updated to match the predictions of the teacher, those that are not sampled remain divergent from the teacher. This means that the unexplored operations would yield high KL divergence from the outputs of the teacher. Therefore, by minimizing the $-\alpha L_{KL}$ term, the $\phi$ variable is updated to encourage the sampling of operations that yield high KL-divergence loss — typically, those that are yet to be explored.
\par
Once the gate values are set for a certain iteration, the function $f_{\theta}^{l,act}$ that maps the inputs of the $l^{th}$ supernet layer to the outputs is computed based on Eq. \ref{eq:map_func}. The $\theta$ denotes the trainable parameters of the candidate operations.
\begin{algorithm}[!b]
\footnotesize
\caption{GTN Training}
\begin{algorithmic}[1]
\STATE \textbf{INPUT:} Teacher $T$ parameterized by $\theta_{T}$, student branches $S = \{s_1, s_2, \ldots, s_{n}\}$ parameterized by $\theta_{S} = \{\theta_{s_1}, \theta_{s_2}, \ldots, \theta_{s_{n}}\}$, and gate params $\Phi = \{\phi^{s_{1},1}, \ldots, \phi^{s_{i},l}, \ldots, \phi^{s_{n},\mathbb{L}_{s_{n}}}\}$
\STATE \textbf{for} {iteration $i$ \textbf{in} total \# of iterations}
    \STATE \quad \textbf{for} {\textbf{all} $s_i$ \textbf{in} $S$}
        \STATE \quad \quad\textbf{for} {layer $l$ in set of layers in $s_i$ 
        }
            \STATE \quad\quad\quad $j_{act}^{l} \sim \phi^{s_{i},l}$
            \STATE \quad\quad\quad set $g^{l,j}$ using $j_{act}^{l}$ based on Eq. \ref{eq:gate}
            \STATE \quad\quad\quad determine $f_{\theta_{s_{i}}}^{l, \text{act}}$ based on Eq. \ref{eq:map_func}
        \STATE \quad\quad \textbf{end for}
    \STATE \quad \textbf{end for}
    \STATE \quad $\hat{y}_{S_{i}} \gets f_{\theta_{S_{i}}}^{\mathbb{L},act} \circ \ldots \circ f_{\theta_{S_{i}}}^{2,act} \circ f_{\theta_{S_{i}}}^{1,act}(x)$
    \STATE \quad \textbf{if} $i \% 2 == 0$ \textbf{then}
        \STATE \quad\quad $optimizer.step(backward(L_{CT}),\theta_{T}, \theta_{S})$
    \STATE \quad \textbf{else}
        \STATE \quad\quad $optimizer.step(backward(L_{\phi}),\Phi)$
    \STATE \quad \textbf{end if}
    \STATE \textbf{end for}
\end{algorithmic}
\label{alg:GTN}
\end{algorithm}
\begin{equation}
    f_{\theta}^{l,act}(x) = \sum_{j}g^{l,j}f_{\theta}^{l,j}(x) \quad \text{where } g^{l,j} \in \{0,1\}
    \label{eq:map_func}
\end{equation}
Using this definition, the 
forward process of each student branch $s_{i}$, containing $\mathbb{L}_{s_{i}}$ number of layers, can be represented by the composite function given in Eq. \ref{eq:inference_func}.
\begin{equation}
\hat{y}_{s_{i}} = f_{\theta_{s_{i}}}^{\mathbb{L}_{s_{i}},act} \circ f_{\theta_{s_{i}}}^{\mathbb{L}_{s_{i}}-1,act} \circ \ldots \circ f_{\theta_{s_{i}}}^{2,act} \circ f_{\theta_{s_{i}}}^{1,act}(x)
\label{eq:inference_func}
\end{equation}
\par
In training our GTN with the supernet architecture, we follow Algorithm \ref{alg:GTN}. The parameters of the sampled operations in the supernet ($\theta_{S}$) and the teacher ($\theta_{T}$) are updated together (line 12). The $\phi$ values, controlling the sampling probability, are updated alternately with $\theta_{S}$ and $\theta_{T}$ (lines 11-15). Initially, $\phi$ is fixed, and operations are sampled for each layer (lines 4-8), which helps regularize the teacher's optimization by minimizing $L_{CT}$. In the following iteration, while $\theta_{S}$ and $\theta_{T}$ are fixed, $\phi$ is updated to minimize $L_{\phi}$ (line 14). This alternating process is repeated until the training concludes after the designated number of epochs.
\subsection{Knowledge Distillation}
After training the GTN, the auxiliary branches containing supernet blocks are discarded and the remaining teacher model can be used to enhance the accuracy of any student model from the pool. At this stage, the target student model is trained with the combined guidance of the ground truth labels and the teacher's logit scores. While any KD optimization objective can be used, we experimented with the vanilla loss function given in Eq. \ref{eq:l_KD} and the one introduced in DKD.
\begin{equation}\belowdisplayskip=0pt
    L_{KD} = L_{CE_{S}} + \alpha L_{KL} \label{eq:l_KD}
\end{equation}
\setlength{\textfloatsep}{10pt}

\begin{table*}[!b]
\centering
\resizebox{1\textwidth}{!}{
\centering
\begin{tabular}{lcccccccccccccccc}
\hline
Dataset                                                                        & \multicolumn{8}{c}{CIFAR-100}                                                                                                       & \multicolumn{8}{c}{ImageNet-200}                                                                       \\ \hline
\multicolumn{1}{l|}{Teacher}                                                   & \multicolumn{4}{c|}{ResNet-32}                                   & \multicolumn{4}{c|}{WRN40-2}                                     & \multicolumn{4}{c|}{ResNet-32}                                   & \multicolumn{4}{c}{EfficientNet-b0} \\ \hline
\multicolumn{1}{l|}{\begin{tabular}[c]{@{}l@{}}Training\\ Method\end{tabular}} & DKD           & SCKD & SFTN & \multicolumn{1}{c|}{GTN (ours)}    & DKD  & SCKD & SFTN          & \multicolumn{1}{c|}{GTN (ours)}    & DKD  & SCKD          & SFTN & \multicolumn{1}{c|}{GTN (ours)}    & DKD   & SCKD & SFTN & GTN (ours)    \\ \hline
\multicolumn{1}{l|}{$\mu_{\Delta}$ ($\uparrow$)}                                 & 0.68          & 0.59 & 2.14 & \multicolumn{1}{c|}{\textbf{2.66}} & 0.46 & 0.47 & 1.77 & \multicolumn{1}{c|}{\textbf{1.99}} & 2.10 & 1.93          & 3.29 & \multicolumn{1}{c|}{\textbf{3.91}} & 4.11  & 3.84 & 4.45 & \textbf{4.82} \\
\multicolumn{1}{l|}{$\sigma_{\Delta}$ ($\downarrow$)}                            & \textbf{0.19} & 0.34 & 0.89 & \multicolumn{1}{c|}{0.38}          & 0.31 & 0.43 & \textbf{0.28} & \multicolumn{1}{c|}{0.38}          & 0.31 & \textbf{0.22} & 0.41 & \multicolumn{1}{c|}{0.33}          & 0.49  & 0.45 & 0.42 & \textbf{0.42} \\ \hline
\end{tabular}
}
\vspace{5pt}
\caption{
Statistical comparison of different KD methods in terms of improvement w.r.t. vanilla KD. $\Delta$ is the relative accuracy improvement (\%) w.r.t. vanilla teachers.  $\mu_{\Delta}$, $\sigma_{\Delta}$ and $\text{range}_{\Delta}$ represent mean, standard deviation, and range of $\Delta$ for KD across 7 different students sampled from the architecture pool. DKD is used as the base distillation method.
}
\label{tab:random_stats}
\end{table*}
\vspace{-10pt}
\section{Experimental Evaluation}
To assess the effectiveness of our method, we evaluate the accuracies of student models randomly selected from a pool of architectures. These students are trained using our method and compared against five different approaches: SFTN, SCKD, DKD, Vanilla KD \cite{hinton2015distilling}, and supervised training (denoted as no-KD). Vanilla KD and DKD, which do not address the capacity gap, establish the lower performance bounds for our comparison. In contrast, SFTN and SCKD, which consider this gap, serve as the main baselines to assess our method's effectiveness in mitigating it. We also include students derived from NAS using ProxylessNAS \cite{cai2018proxyless}, comparing their performance on CIFAR-100 \cite{krizhevsky2009learning} and ImageNet-200 \cite{deng2009imagenet}, to further validate our approach. Lastly, we display the memory sizes of these student models in conjunction with the on-chip memory availability of three edge devices to showcase a real-world scenario of tailoring student models for specific hardware platforms. Our code is available at \url{https://github.com/kuluhan/GTN}.
\vspace{-10pt}
\paragraph{Implementation Details:} In our KD experiments, we employ three teacher architectures: ResNet-32, WRN40-2, and EfficientNet-b0 \cite{he2016deep, zagoruyko2016wide, tan2019efficientnet}. 
During teacher training, our GTN framework modifies the teacher architecture by grafting blocks of the reference supernet architecture, onto the teacher to form different student branches. Once the teacher is trained, these branches are discarded and the teacher model is used to train student models via KD. The supernet is constructed with ResNet layers varying in depth and filter size, equipped with identity and zero operations to allow flexible architecture sampling. For knowledge transfer, we use Vanilla KD and DKD methods, training student models over 240 epochs with learning rates of $0.05$ and $0.1$ for CIFAR-100 and ImageNet-200 datasets respectively. Cosine annealing is used for adjusting the learning rates. Further implementation details and training configurations are provided in the Appendix.
\vspace{-5pt}
\begin{table*}[!t]
\centering
\resizebox{0.82\textwidth}{!}{
\begin{tabular}{ccccccccc}
\hline
\multicolumn{1}{c|}{Method}                                                      & \multicolumn{1}{c|}{no KD}                       & \multicolumn{1}{c|}{Vanilla KD}                    & \multicolumn{1}{c|}{SCKD}                          & \multicolumn{4}{c|}{SFTN}                                                                                                                                                                                      & GTN (ours)                                       \\ \hline
\multicolumn{9}{c}{\cellcolor[HTML]{C0C0C0}CIFAR-100}                                                                                                                                                                                                                                                                                                                                                                                                                                                             \\ \hline
\multicolumn{1}{c|}{\begin{tabular}[c]{@{}c@{}}Reference\\ Student\end{tabular}} & \multicolumn{1}{c|}{N/A}                         & \multicolumn{1}{c|}{N/A}                           & \multicolumn{1}{c|}{N/A}                           & \multicolumn{1}{c|}{s1}                            & \multicolumn{1}{c|}{s2}                         & \multicolumn{1}{c|}{s3}                            & \multicolumn{1}{c|}{s4}                            & supernet                                         \\ \hline
\multicolumn{1}{c|}{Teacher acc.}                                                & \multicolumn{1}{c|}{N/A}                         & \multicolumn{1}{c|}{78.04}                         & \multicolumn{1}{c|}{78.04}                         & \multicolumn{1}{c|}{81.54}                         & \multicolumn{1}{c|}{81.46}                      & \multicolumn{1}{c|}{81.49}                         & \multicolumn{1}{c|}{80.97}                         & 80.71                                            \\ \hline
\multicolumn{1}{c|}{s1 acc. ($\Delta$)}                                          & \multicolumn{1}{c|}{73.22}                       & \multicolumn{1}{c|}{75.40}                         & \multicolumn{1}{c|}{75.34 ($-0.06$)}               & \multicolumn{1}{c|}{75.82 ($+0.42$)}               & \multicolumn{1}{c|}{76.09 ($+0.69$)}            & \multicolumn{1}{c|}{75.34 ($-0.06$)}               & \multicolumn{1}{c|}{76.03 ($+0.63$)}               & \textbf{76.70 ($+1.30$)}                         \\
\multicolumn{1}{c|}{s2 acc. ($\Delta$)}                                          & \multicolumn{1}{c|}{77.92}                       & \multicolumn{1}{c|}{77.89}                         & \multicolumn{1}{c|}{77.47 ($-0.42$)}               & \multicolumn{1}{c|}{78.23 ($+0.34$)}               & \multicolumn{1}{c|}{78.52 ($+0.63$)}            & \multicolumn{1}{c|}{77.90 ($+0.01$)}               & \multicolumn{1}{c|}{77.76 ($-0.13$)}               & \textbf{78.85 ($+0.96$)}                         \\
\multicolumn{1}{c|}{s3 acc. ($\Delta$)}                                          & \multicolumn{1}{c|}{76.87}                       & \multicolumn{1}{c|}{77.21}                         & \multicolumn{1}{c|}{77.08 ($-0.13$)}               & \multicolumn{1}{c|}{77.66 ($+0.45$)}               & \multicolumn{1}{c|}{78.05 ($+0.84$)}            & \multicolumn{1}{c|}{78.05 ($+0.84$)}               & \multicolumn{1}{c|}{77.94 ($+0.73$)}               & \textbf{78.22 ($+1.01$)}                         \\
\multicolumn{1}{c|}{s4 acc. ($\Delta$)}                                          & \multicolumn{1}{c|}{75.60}                       & \multicolumn{1}{c|}{76.42}                         & \multicolumn{1}{c|}{75.77 ($-0.65$)}               & \multicolumn{1}{c|}{77.20 ($+0.78$)}               & \multicolumn{1}{c|}{77.71 ($+1.29$)}            & \multicolumn{1}{c|}{77.23 ($+0.81$)}               & \multicolumn{1}{c|}{77.66 ($+1.24$)}               & \textbf{77.79 ($+1.37$)}                         \\ \hline
\multicolumn{9}{c}{\cellcolor[HTML]{C0C0C0}ImageNet-200}                                                                                                                                                                                                                                                                                                                                                                                                                                                          \\ \hline
\multicolumn{1}{c|}{\begin{tabular}[c]{@{}c@{}}Reference\\ Student\end{tabular}} & \multicolumn{1}{c|}{\cellcolor[HTML]{FFFFFF}N/A} & \multicolumn{1}{c|}{\cellcolor[HTML]{FFFFFF}N/A}   & \multicolumn{1}{c|}{\cellcolor[HTML]{FFFFFF}N/A}   & \multicolumn{1}{c|}{\cellcolor[HTML]{FFFFFF}s5}    & \multicolumn{1}{c|}{\cellcolor[HTML]{FFFFFF}s6} & \multicolumn{1}{c|}{\cellcolor[HTML]{FFFFFF}s7}    & \multicolumn{1}{c|}{\cellcolor[HTML]{FFFFFF}s8}    & supernet                                         \\ \hline
\multicolumn{1}{c|}{Teacher acc.}                                                & \multicolumn{1}{c|}{\cellcolor[HTML]{FFFFFF}N/A} & \multicolumn{1}{c|}{\cellcolor[HTML]{FFFFFF}65.28} & \multicolumn{1}{c|}{\cellcolor[HTML]{FFFFFF}65.28} & \multicolumn{1}{c|}{\cellcolor[HTML]{FFFFFF}69.71} & \multicolumn{1}{c|}{69.08}                      & \multicolumn{1}{c|}{\cellcolor[HTML]{FFFFFF}68.96} & \multicolumn{1}{c|}{\cellcolor[HTML]{FFFFFF}69.00} & \cellcolor[HTML]{FFFFFF}70.10                    \\ \hline
\multicolumn{1}{c|}{s5 acc. ($\Delta$)}                                          & \multicolumn{1}{c|}{62.63}                       & \multicolumn{1}{c|}{64.86}                         & \multicolumn{1}{c|}{65.36 ($+0.50$)}               & \multicolumn{1}{c|}{65.54 ($+0.68$)}               & \multicolumn{1}{c|}{64.99 ($+0.13$)}            & \multicolumn{1}{c|}{65.08 ($+0.22$)}               & \multicolumn{1}{c|}{65.30 ($+0.44$)}               & \cellcolor[HTML]{FFFFFF}\textbf{65.63 ($+0.77$)} \\
\multicolumn{1}{c|}{s6 acc. ($\Delta$)}                                          & \multicolumn{1}{c|}{64.34}                       & \multicolumn{1}{c|}{64.79}                         & \multicolumn{1}{c|}{65.68 ($+0.45$)}               & \multicolumn{1}{c|}{66.00 ($+1.21$)}               & \multicolumn{1}{c|}{65.71 ($+0.92$)}            & \multicolumn{1}{c|}{65.88 ($+1.09$)}               & \multicolumn{1}{c|}{65.62 ($+0.83$)}               & \textbf{66.08 ($+1.29$)}                         \\
\multicolumn{1}{c|}{s7 acc. ($\Delta$)}                                          & \multicolumn{1}{c|}{62.62}                       & \multicolumn{1}{c|}{64.37}                         & \multicolumn{1}{c|}{63.97 ($-0.40$)}               & \multicolumn{1}{c|}{64.69 ($+0.32$)}               & \multicolumn{1}{c|}{64.73 ($+0.36$)}            & \multicolumn{1}{c|}{64.71 ($+0.34$)}               & \multicolumn{1}{c|}{65.13 ($+0.76$)}               & \textbf{65.74 ($+1.37$)}                         \\
\multicolumn{1}{c|}{s8 acc. ($\Delta$)}                                          & \multicolumn{1}{c|}{62.15}                       & \multicolumn{1}{c|}{63.53}                         & \multicolumn{1}{c|}{63.95 ($+0.42$)}               & \multicolumn{1}{c|}{63.48 ($-0.05$)}               & \multicolumn{1}{c|}{64.13 ($+0.60$)}            & \multicolumn{1}{c|}{63.88 ($+0.35$)}               & \multicolumn{1}{c|}{64.13 ($+0.60$)}               & \textbf{64.29 ($+0.76$)}                         \\ \hline
\end{tabular}
}
\vspace{5pt}
\caption{Accuracy values (\%) of randomly selected student models distilled using different methods. $\Delta$ denotes improvement (\%) over vanilla KD. Vanilla KD is used as the base distillation method.}
\label{tab:random_CIFAR100}
\vspace{-5pt}
\end{table*}
\vspace{-5pt}
\paragraph{KD with random students:}  We first compare our KD approach with baseline methods based on the performance of student models randomly drawn from the pool of architectures represented by the supernet. For each dataset, we randomly select seven student architectures. Each method except SFTN uses a single teacher model leading to seven teacher-student pairings for each dataset and teacher architecture combination (e.g. ResNet-32 \& CIFAR-100). As for SFTN, we train four teachers, each specialized for a different student, leading to twenty-eight teacher-student pairings. The results are summarized in Table \ref{tab:random_stats}, with $\Delta$ indicating the relative improvements compared to the vanilla KD method. $\mu_{\Delta}$ and $\sigma_{\Delta}$ represent the mean and standard deviation of $\Delta$ recorded across seven different reference students. The $\text{range}_{\Delta}$ is the gap between the maximum and minimum $\Delta$. The results showcase GTN's consistent advantage in improving student accuracy, underscoring its versatility in mitigating the teacher-student capacity gap. Table \ref{tab:random_CIFAR100} details individual accuracies for four students among the aforementioned seven randomly sampled students per dataset, trained with the supervision of ResNet-32 teachers. These are identified as $s1, s2, s3, s4$ in CIFAR-100 and $s5, s6, s7, s8$ in ImageNet-200. While GTN generally enhances KD accuracy across all experiments SFTN sometimes result in accuracy decreases when applied to non-reference students, as noted in Table \ref{tab:random_CIFAR100}. For instance, a teacher optimized for $s4$ might yield reduced accuracy, i.e. $\Delta = -0.13$ when used to train $s2$, indicating that SFTN training does not universally benefit all student types. SCKD also causes accuracy drops for some students. Additionaly, our findings reveal that SFTN does not always yield the highest accuracy improvements for the students it is specialized for. For example, a teacher trained specifically for $s2$ might achieve a larger improvement for $s4$,  achieving a $\Delta$ of $+1.29$ vs $+0.63$, suggesting that while SFTN acts as a regularizer, it may inadvertently benefit other students more than the intended ones. SFTN and GTN teachers generally achieve higher accuracy than vanilla teachers on target test sets due to regularization process they involve preventing overfitting. However, as seen in Table \ref{tab:random_CIFAR100}, teacher model accuracy does not directly correlate with effectiveness in KD.
\par
\begin{figure*}[!b]
\vspace{-5pt}
  \centering
    \includegraphics[width=0.80\textwidth]{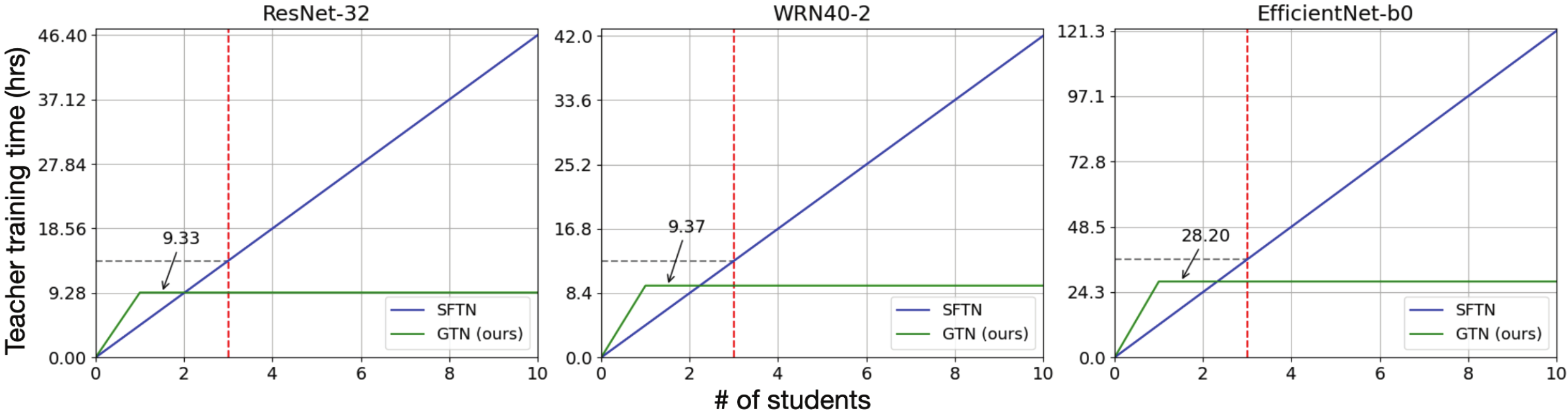}
    \vspace{5pt}
   \caption{Training time comparison for ResNet32, WRN40-2 and EfficientNet-b0 teachers respectively. Dashed vertical lines colored in red mark the \# of students after which our GTN method attains the time cost advantage.}
   \label{fig:time_cost}
\end{figure*}

While consistently improving accuracy across student models, our method incurs only a one-off time cost that is slightly under that of training three SFTN teachers. As seen in Fig. \ref{fig:time_cost} the time cost of SFTN scales linearly with the number of students involved. When more than two deployment scenarios exist, our GTN approach takes over the time advantage.
\vspace{-12pt}
\paragraph{KD with students obtained by NAS} Besides selecting the student models for different deployment scenarios manually, NAS can also be employed to sample students that exhibit close to optimal performance within the given resource budgets. To test the effectiveness of our method in improving KD accuracy for NAS-discovered student architectures, we search our pool of student architectures to obtain three distinct student architectures of varying sizes. Since we represent the architecture pool using a supernet, we employ a differentiable search strategy, that is ProxylessNAS \cite{cai2018proxyless}. To obtain each student, we limit the searched model size by a different number of layers. The largest architecture, denoted as $s_{nas}^{l}$, is derived from the full-scale supernet consisting of 6 searchable layers without any constraints. For the medium ($s_{nas}^{m}$) and small ($s_{nas}^{s}$) architectures, we reduce the number of searchable layers to 5 and 3, respectively. Subsequently, to prepare the SFTN baseline for comparison, we train specialized teachers using these three architectures as reference students. We then evaluate these teachers based on their KD performance for all three NAS-discovered students. We again use DKD as the base distillation method in this evaluation. From Table \ref{tab:nas_results} we can observe that GTN method again caused the largest improvement in student accuracies in almost all experiments. This means that when a new student is configured via NAS due to changes in the deployment platform, our GTN method is more likely to provide higher accuracy benefits through KD. 
\begin{table*}[!t]
\vspace{-5pt}
\centering
\resizebox{1\textwidth}{!}{
\begin{tabular}{lcccccccccccccccc}
\hline
Dataset                                                                        & \multicolumn{8}{c}{CIFAR-100}                                                                                             & \multicolumn{8}{c}{ImageNet-200}                                                                       \\ \hline
\multicolumn{1}{l|}{Teacher}                                                   & \multicolumn{4}{c|}{ResNet-32}                              & \multicolumn{4}{c|}{WRN40-2}                                & \multicolumn{4}{c|}{ResNet-32}                                & \multicolumn{4}{c}{EfficientNet-b0}    \\ \hline
\multicolumn{1}{l|}{\begin{tabular}[c]{@{}l@{}}Training\\ Method\end{tabular}} & DKD   & SCKD  & SFTN  & \multicolumn{1}{c|}{GTN (ours)}     & DKD   & SCKD  & SFTN  & \multicolumn{1}{c|}{GTN (ours)}     & DKD   & SCKD  & SFTN    & \multicolumn{1}{c|}{GTN (ours)}     & DKD   & SCKD  & SFTN  & GTN (ours)     \\ \hline
\multicolumn{1}{c|}{$s_{nas}^{l}$ acc.}                                          & 78.18 & 77.51 & 79.48 & \multicolumn{1}{c|}{\textbf{79.95}} & 77.98 & 78.02 & 79.33 & \multicolumn{1}{c|}{\textbf{79.55}} & 67.09 & 66.17 & 68.17 & \multicolumn{1}{c|}{\textbf{68.87}} & 69.63 & 69.26 & 69.78 & \textbf{70.19} \\
\multicolumn{1}{c|}{$s_{nas}^{m}$ acc.}                                          & 77.64 & 77.89 & 79.35 & \multicolumn{1}{c|}{\textbf{79.63}} & 77.96 & 78.31 & 79.27 & \multicolumn{1}{c|}{\textbf{79.33}} & 65.67 & 65.97 & 67.24  & \multicolumn{1}{c|}{\textbf{67.41}} & 67.83 & 67.77 & 68.44 & \textbf{68.48} \\
\multicolumn{1}{c|}{$s_{nas}^{s}$ acc.}                                          & 75.76 & 76.08 & 76.16 & \multicolumn{1}{c|}{\textbf{77.23}} & 75.77 & 75.79 & 76.45 & \multicolumn{1}{c|}{\textbf{76.80}} & 65.19 & 65.35 & 66.63   & \multicolumn{1}{c|}{\textbf{67.52}} & 67.42 & 67.27 & 67.92 & \textbf{68.55} \\ \hline
\end{tabular}
}
\vspace{5pt}
\caption{Accuracy values (\%) of student models obtained by NAS, distilled using different methods. $\Delta$ denotes improvement (\%) over vanilla KD. DKD is used as the main distillation method.}
\label{tab:nas_results}
\vspace{-10pt}
\end{table*}
\par
In Table \ref{tab:edge-device-compat}, we display the memory sizes of seven student models we considered in CIFAR-100 experiments, detailing both the full-scale (32-bit) and the 8-bit weight quantized versions. The varied memory sizes of these student models, demonstrate the diversity of our student model pool, which contains a range of architectures with different capacities. Moreover, we provide the on-chip memory capacities of three edge devices—Arm Ethos N77 \cite{arm2020} , Edge TPU \cite{google2020}, and Raspberry Pi \cite{raspberrypi2024}—, focusing on the feasibility of deploying each student model on these platforms.  The green checkmarks and red Xs, show whether each 8-bit quantized model fits within the memory constraints of these devices. 
\vspace{-15pt}
\begin{table}[!t]
\centering
\resizebox{0.8\textwidth}{!}{%
\centering
\begin{tabular}{@{}lcccccccccccccc@{}}
\toprule
          & \multicolumn{2}{c}{\textbf{s1}} & \multicolumn{2}{c}{\textbf{s2}} & \multicolumn{2}{c}{\textbf{s3}} & \multicolumn{2}{c}{\textbf{s4}} & \multicolumn{2}{c}{\textbf{s\textsuperscript{s}\textsubscript{nas}}} & \multicolumn{2}{c}{\textbf{s\textsuperscript{m}\textsubscript{nas}}} & \multicolumn{2}{c}{\textbf{s\textsuperscript{l}\textsubscript{nas}}} \\ \cmidrule{2-15}
\textbf{Bit-width}          & 32b   & 8b    & 32b   & 8b    & 32b   & 8b    & 32b   & 8b    & 32b   & 8b   & 32b   & 8b   & 32b & 8b \\ 
\midrule
 \textbf{Memory Size (MB)}                          & 22.2   & 5.5    & 29.9   & 7.5    & 50.6   & 12.6    & 25.4   & 6.4    & 6.7   & 1.7   & 28.6   & 7.2   & 77.1 & 19.3     \\ \midrule
Arm Ethos N77 (1-4 MB)    & \multicolumn{2}{c}{\cellcolor{red!25}X}      & \multicolumn{2}{c}{\cellcolor{red!25}X}      & \multicolumn{2}{c}{\cellcolor{red!25}X}      & \multicolumn{2}{c}{\cellcolor{red!25}X}      & \multicolumn{2}{c}{\cellcolor{green!25}\checkmark} & \multicolumn{2}{c}{\cellcolor{red!25}X}      & \multicolumn{2}{c}{\cellcolor{red!25}X}      \\
Edge TPU (8 MB)           & \multicolumn{2}{c}{\cellcolor{green!25}\checkmark} & \multicolumn{2}{c}{\cellcolor{green!25}\checkmark} & \multicolumn{2}{c}{\cellcolor{red!25}X}      & \multicolumn{2}{c}{\cellcolor{green!25}\checkmark} & \multicolumn{2}{c}{\cellcolor{green!25}\checkmark} & \multicolumn{2}{c}{\cellcolor{green!25}\checkmark} & \multicolumn{2}{c}{\cellcolor{red!25}X}      \\
Raspberry Pi (1 GB)       & \multicolumn{2}{c}{\cellcolor{green!25}\checkmark} & \multicolumn{2}{c}{\cellcolor{green!25}\checkmark} & \multicolumn{2}{c}{\cellcolor{green!25}\checkmark} & \multicolumn{2}{c}{\cellcolor{green!25}\checkmark} & \multicolumn{2}{c}{\cellcolor{green!25}\checkmark} & \multicolumn{2}{c}{\cellcolor{green!25}\checkmark} & \multicolumn{2}{c}{\cellcolor{green!25}\checkmark}  \\ \bottomrule
\end{tabular}%
}
\vspace{5pt}
\caption{Compatibility of Edge Devices with Various Student Models}
\label{tab:edge-device-compat}
\vspace{-5pt}
\end{table}
\section{Conclusion}
\vspace{-5pt}
 In conclusion, this study tackles the teacher-student capacity gap problem, which undermines the effectiveness of Knowledge Distillation (KD) in optimizing neural networks. Previous methods have concentrated on customizing teacher-student pairs, a process that needs to be repeated for each student architecture requiring KD. This characteristic renders these approaches impractical, particularly in scenarios involving multiple deployment platforms, each demanding KD for distinct deployment-ready students. Our novel approach introduces a single generic teacher model capable of transferring knowledge effectively across a variety of student architectures. The proposed technique's advantage is that it both improves KD performance and maintains constant time cost, which is equivalent to that of a few specialized teachers. Moreover, the adaptability of our approach to NAS scenarios further highlights its practicality.

\section{Acknowledgement}
This work is supported by the National Research Foundation, Singapore under its Competitive Research Programme Award NRF-CRP23-2019-0003 and Singapore Ministry of Education Academic Research Fund T1 251RES1905. We thank Cihan Acar and Liu Siying for helpful comments on the manuscript.

\nocite{eyono2021autokd}

\bibliography{egbib}
\end{document}